%% file: main.tex
\documentclass[10pt,twocolumn,letterpaper]{article}

\usepackage{cvpr}              % To produce the CAMERA-READY version
\usepackage[accsupp]{axessibility} 
\input{preamble}

\definecolor{cvprblue}{rgb}{0.21,0.49,0.74}
\usepackage[pagebackref,breaklinks,colorlinks,allcolors=cvprblue]{hyperref}

\def\confName{CVPR}
\def\confYear{2026}

\title{OmniDrive-R1: Reinforcement-driven Interleaved Multi-modal Chain-of-Thought for Trustworthy Vision-Language Autonomous Driving}

\author{ 
    Zhenguo Zhang\textsuperscript{1}\thanks{Equal contribution.}, 
    Haohan Zheng\textsuperscript{2}\footnotemark[1], 
    Yishen Wang\textsuperscript{3}, 
    Le Xu\textsuperscript{5,6}, 
    Tianchen Deng\textsuperscript{4}\thanks{Corresponding author.}, \\
    Xuefeng Chen\textsuperscript{2}, 
    Qu Chen\textsuperscript{5,6}, 
    Bo Zhang\textsuperscript{5,6}, 
    Wuxiong Huang\textsuperscript{5,6}
    \\
    {\textsuperscript{1} ShanghaiTech University} \quad
    {\textsuperscript{2} Tsinghua University} \quad
    {\textsuperscript{3} Tongji University} \\
    {\textsuperscript{4} Shanghai Jiao Tong University} \quad
    {\textsuperscript{5} MEGVII Technology} \quad
    {\textsuperscript{6} AFARI}
}

\begin{document}
\maketitle
\input{sec/0_abstract}    
\input{sec/1_intro}

\input{sec/2_relatedwork}
\input{sec/3_method}
\input{sec/4_exp}

\input{sec/5_conclusion}

{
    \small
    \bibliographystyle{ieeenat_fullname}
    \bibliography{main}
}

% WARNING: do not forget to delete the supplementary pages from your submission 
\input{sec/X_suppl}

\end{document}

%% file: preamble.tex
\usepackage{graphicx}	
\usepackage{amsmath}	
\usepackage{amssymb}	
\usepackage{booktabs}
\usepackage{times}
\usepackage{microtype}
\usepackage{epsfig}
\usepackage{caption}
\usepackage{float}
\usepackage{placeins}
\usepackage{color, colortbl}
\usepackage{stfloats}
\usepackage{enumitem}
\usepackage{tabularx}
\usepackage{xstring}
\usepackage{multirow}
\usepackage{xspace}
\usepackage{url}
\usepackage{subcaption}
\usepackage[pagebackref,breaklinks,colorlinks,allcolors=cvprblue]{hyperref}

\usepackage[markup=underlined,deletedmarkup=sout,color=blue]{changes}

\usepackage[capitalize]{cleveref}
\crefname{section}{Sec.}{Secs.}
\Crefname{section}{Section}{Sections}
\Crefname{table}{Table}{Tables}
\crefname{table}{Tab.}{Tabs.}

% ...

\usepackage{xcolor}

\usepackage{pifont} % For \ding commands (check/x marks)
\newcommand{\mycmark}{\ding{52}} % Checkmark
\newcommand{\myxmark}{\ding{55}} % X-mark

%% file: sec/0_abstract.tex
\begin{abstract}
The deployment of Vision-Language Models (VLMs) in safety-critical domains like autonomous driving (AD) is critically hindered by reliability failures, most notably object hallucination. This failure stems from their reliance on ungrounded, text-based Chain-of-Thought (CoT) reasoning. While existing multi-modal CoT approaches attempt mitigation, they suffer from two fundamental flaws: (1) decoupled perception and reasoning stages that prevent end-to-end joint optimization, and (2)  reliance on expensive, dense localization labels. Thus we introduce OmniDrive-R1, an end-to-end VLM framework designed for autonomous driving, which unifies perception and reasoning through an interleaved Multi-modal Chain-of-Thought (iMCoT) mechanism. Our core innovation is an Reinforcement-driven visual grounding capability, enabling the model to autonomously direct its attention and "zoom in" on critical regions for fine-grained analysis. This capability is enabled by our pure two-stage reinforcement learning training pipeline and Clip-GRPO algorithm. Crucially, Clip-GRPO introduces an annotation-free, process-based grounding reward. This reward not only eliminates the need for dense labels but also circumvents the instability of external tool calls by enforcing real-time cross-modal consistency between the visual focus and the textual reasoning. Extensive experiments on DriveLMM-o1 demonstrate our model's significant improvements. Compared to the baseline Qwen2.5VL-7B, OmniDrive-R1 improves the overall reasoning score from 51.77\% to 80.35\%, and the final answer accuracy from 37.81\% to 73.62\%.  
\end{abstract}

%% file: sec/1_intro.tex
\section{Introduction}
\label{sec:intro}
\begin{figure*}[t]
    \centering
    \includegraphics[width=\textwidth]{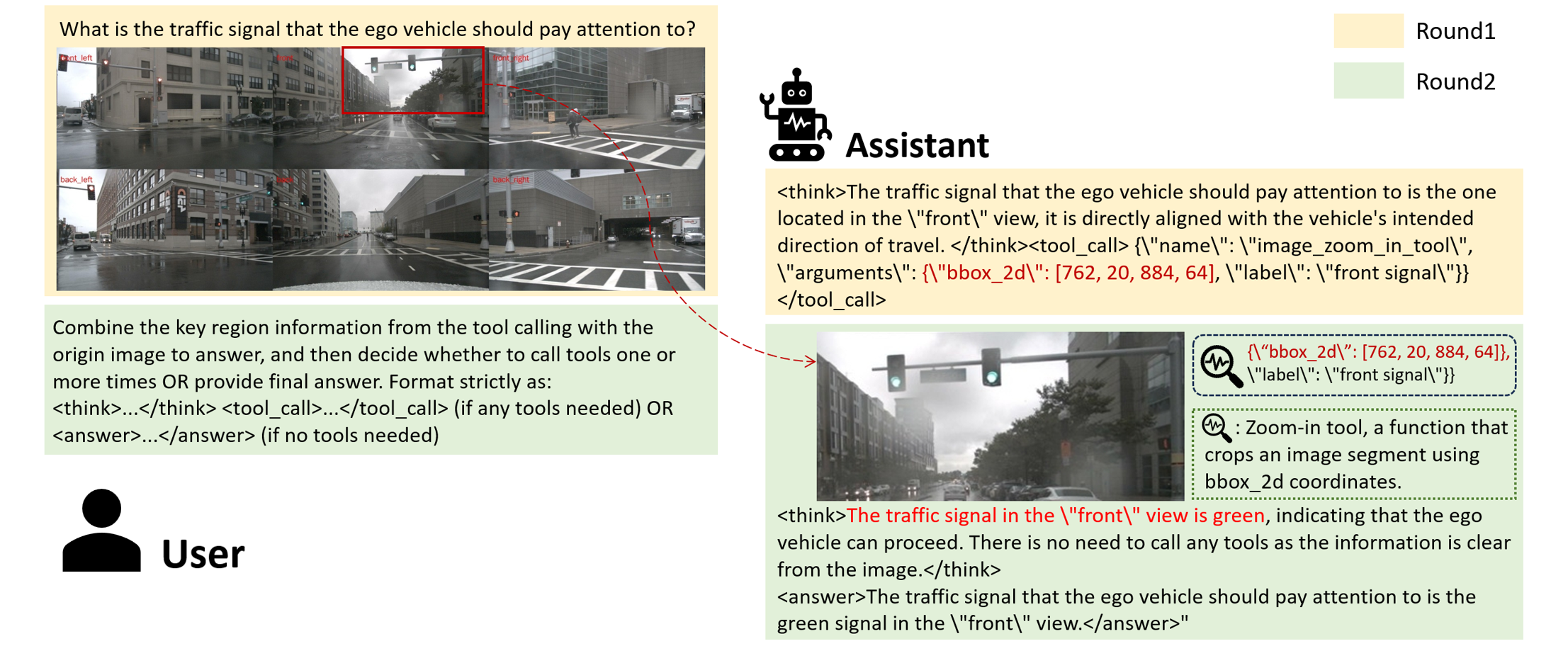}
    \caption{\textbf{An Illustration of OmniDrive-R1’s Interleaved Multi-modal Chain-of-Thought Reasoning Example.} The model initiates a multi-step thought process (Round 1) by actively invoking the Zoom-in Tool to ground its reasoning on a critical region (the traffic signal). This mechanism dynamically acquires fine-grained visual evidence (Round 2), which is directly used to refine the thought and arrive at a confident, visually-backed final answer. This active, evidence-based process significantly enhances grounding and interpretability.}
    \label{fig:fig0}
\end{figure*}
The advancement of autonomous driving (AD) systems has fundamentally shifted from pure object detection and tracking to intricate, high-level reasoning and decision-making under complex, safety-critical scenarios \cite{wang2025omnidrive, sima2023driving}.  Meanwhile, broader progress in scene representation and collaborative perception continues to reinforce the need for richer environment understanding, providing an important backdrop for the development of more capable autonomous driving systems \cite{deng2025best3dscenerepresentation, deng2025mne, deng2025mcnslammultiagentcollaborativeneural}.Tasks such as intent prediction, causal explanation, and abstract policy planning require capabilities beyond simple perception, making them essential for robust driving deployment \cite{qian2025agentthink, ishaq2025drivelmm}. Vision-Language Models (VLMs), which seamlessly integrate visual perception with powerful linguistic reasoning, have emerged as a promising foundation for such complex, cognitive AD agents \cite{xu2025llava}. By leveraging Chain-of-Thought (CoT) reasoning \cite{wei2022chain}, VLMs can articulate their decision-making process, offering transparency and interpretability.

Despite this potential, the deployment of VLMs in AD is critically hindered by fundamental reliability failures, most notably object hallucination \cite{an2025mitigating, zheng2025modality}. This failure is rooted in the current reliance on ungrounded, text-based CoT reasoning. When the model’s "thought process" is largely confined to discrete textual tokens, it risks losing fidelity to the continuous, dynamic visual scene, leading to the fabrication of non-existent objects, states, or relationships. In autonomous driving, a model that hallucinates even a minor detail can lead to catastrophic and life-threatening failures.

To mitigate this challenge, a new paradigm of multi-modal interactive reasoning is emerging \cite{li2025imagine,cheng2025visual,hu2024visual}. Recent works have proposed predefined workflow based strategies \cite{shao2024visual,sun2024visual,jiang2025vlm} and tool-augmented methods \cite{qian2025agentthink} to incorporate visual information into CoT reasoning. The former often employs rigid, hard-coded reasoning paths. This structural inflexibility severely limits their adaptability, confining them to specific, pre-defined problem types and failing to generalize to novel scenarios. The latter delegates critical perception sub-tasks to a collection of disparate, external models. This decoupled architecture fundamentally compromises the integrity of the perception-reasoning process, preventing end-to-end joint optimization and making it difficult to ensure cross-modal consistency. Consequently, these methods not only yield suboptimal solutions \cite{ross2011reduction} but also fail to cultivate and leverage the intrinsic, fine-grained visual processing potential of the core VLM itself. Critically, both paradigms are further constrained by their reliance on large-scale, high-quality annotated reasoning data, which is expensive and difficult to acquire.

In contrast, human drivers primarily rely on a dynamic interplay between their cognitive processes and the visual information from their surroundings to make judgments and take action. This requires a continuous and dynamic interaction with visual data. Inspired by this, we hypothesize that a VLM can achieve the same goal by exclusively optimizing its native perception, and reasoning capabilities. Here, perception is analogous to a driver's cognitive interaction with the scene to acquire visual information, while reasoning corresponds to their process of making sense of the information. 

Based on this insight, we introduce \textbf{OmniDrive-R1}, an end-to-end VLM framework designed for autonomous driving. As illustrated in \cref{fig:fig0}, OmniDrive-R1 empowers the VLM with an adaptive, active perception capability via an interleaved Multi-modal Chain-of-Thought (iMCoT) mechanism. Our core technical innovation is a reinforcement-driven visual grounding capability, enabling the model to autonomously direct its attention and zoom in on critical regions for fine-grained analysis during the reasoning process. This activation is purely driven by the base VLM's intrinsic grounding potential, eliminating reliance on any external tools. Concretely, this capability is realized through a pure two-stage Reinforcement Learning (RL) training strategy utilizing our novel Clip-GRPO. Building upon Group Relative Policy Optimization (GRPO) \cite{shao2024deepseekmath}, Clip-GRPO introduces an innovative, annotation-free, process-based grounding reward. This reward uses the CLIP model’s \cite{radford2021learning} cross-modal consistency to enforce real-time alignment between the model's visual focus and its textual reasoning, thereby eliminating the need for dense localization labels and circumventing the instability of external tool calls. 

In summary, our major contributions are as follows:

\begin{enumerate}
    \item We propose \textbf{OmniDrive-R1}, the first purely RL-driven end to end VLM-based framework for AD. OmniDrive-R1 unifies perception and reasoning through an iMCoT mechanism, which leverages the VLM's native grounding capability to dynamically locate the most task-relevant critical regions for fine-grained analysis, without reliance on external models.

    \item We introduce \textbf{Clip-GRPO}, a novel RL training strategy that addresses the challenges of label dependency and instability in tool learning. It innovatively incorporates an annotation-free, process-based grounding reward based on CLIP's cross-modal consistency, significantly enhancing the coherence between the model's native grounding and textual reasoning.

    \item We demonstrate that OmniDrive-R1 achieves statistically consistent \textbf{state-of-the-art (SOTA)} performance on autonomous driving reasoning benchmarks, significantly surpassing leading industry models (e.g., GPT-4o \cite{islam2024gpt}, Qwen2.5VL-72B \cite{bai2025qwen2}), and SFT/workflow based VLMs (e.g., Agentthink \cite{qian2025agentthink}, DriveLMM-o1 \cite{ishaq2025drivelmm}).
\end{enumerate}

%% file: sec/2_relatedwork.tex
\section{Related Work}
\label{sec:formatting}

\subsection{Multi-modal LLM Reasoning}

The rapid advancement of large language models (LLMs) has provided a strong foundation for multi-modal large language models (MLLMs). 
LLMs exhibit impressive reasoning abilities, largely attributed to CoT prompting \cite{qi2024mutual,ning2023skeleton,wang2022self,lightman2023let,wei2022chain,snell2024scaling}, which guides models to generate interpretable intermediate steps and emulate human reasoning. 
Extending this paradigm to multi-modal contexts, recent research has explored both training-based and prompting-based strategies to enhance MLLM reasoning.

From the training perspective, several works introduce multi-modal reasoning chains \cite{xu2025llavacotletvisionlanguage,thawakar2025llamavo1rethinkingstepbystepvisual,sun2025mm,li2025imagine} and construct high-quality reasoning datasets \cite{du2025virgo,shen2025vlm}. 
LLaVA-CoT \cite{xu2025llavacotletvisionlanguage} employs a four-stage reasoning process with structured annotations, while LIamav-o1 \cite{thawakar2025llamavo1rethinkingstepbystepvisual} integrates curriculum learning and beam search for efficient progression. 
V* \cite{wu2024v} leverages world knowledge for enhanced contextual grounding, and Virgo \cite{du2025virgo} demonstrates that textual reasoning data can effectively trigger “slow-thinking” behaviors in MLLMs, while Hume \cite{song2025hume} further explores System-2 thinking in Vision-Language-Action models. 
Pretraining methods such as Flamingo \cite{alayrac2022flamingo} and KOSMOS-2 \cite{peng2023kosmos} further bridge the modality gap by learning from interleaved or grounded image-text pairs \cite{luo2020multi,zhu2022seqtr}.

From the prompting perspective, studies focus on prompt design to improve understanding of complex visual tasks \cite{zhang2024prompt,zhang2023makes,hong2024cogagent,mitra2024compositional,zheng2023ddcot}. 
For instance, highlighting prompt tokens allows controllable and interactive generation in MLLMs without additional training \cite{zhang2024prompt}. 
However, most approaches remain predominantly text-driven, limiting their effectiveness in vision-intensive reasoning, where visual grounding and step-level interpretability are essential.

\subsection{Multi-modal LLM in Autonomous Driving}

The integration of MLLMs/VLMs has reshaped end-to-end autonomous driving, enabling powerful scene understanding and reasoning \cite{fu2024drive,xu2024drivegpt4,qian2024nuscenes,deng2025gaussiandwm3dgaussiandriving}. 
NuScenes-QA \cite{qian2024nuscenes} established the first VQA benchmark for driving scenarios, and Talk2BEV \cite{choudhary2024talk2bev} incorporated bird’s-eye-view representations for multi-task visual reasoning. 
More recent works \cite{nie2024reason2drive,corbiere2025retrieval, fu2025minddrive} emphasize reasoning enhancement, such as Reason2Drive \cite{nie2024reason2drive}, which introduces a video-text benchmark and an aggregated evaluation metric for chain-based reasoning, and DrivingVQA \cite{corbiere2025retrieval}, which adopts retrieval-based visual CoT reasoning to reduce reliance on textual cues. 
To further mitigate hallucination and inefficiency, AgentThink \cite{qian2025agentthink} integrates CoT reasoning with dynamic tool invocation via supervised fine-tuning and GRPO.

While these efforts advance reasoning outcomes, most methods neglect the fidelity of intermediate reasoning steps, which is critical for safety-critical domains like autonomous driving. 
Ensuring logical consistency and accuracy at every reasoning stage remains an open challenge. 
To this end, we propose a two-stage reinforcement learning framework that explicitly optimizes reasoning reliability and introduces an evaluation protocol to assess step-level accuracy and coherence.

%% file: sec/3_method.tex
\section{Method}

\label{sec:method}
\begin{figure*}[t]
    \centering
    \includegraphics[width=\textwidth]{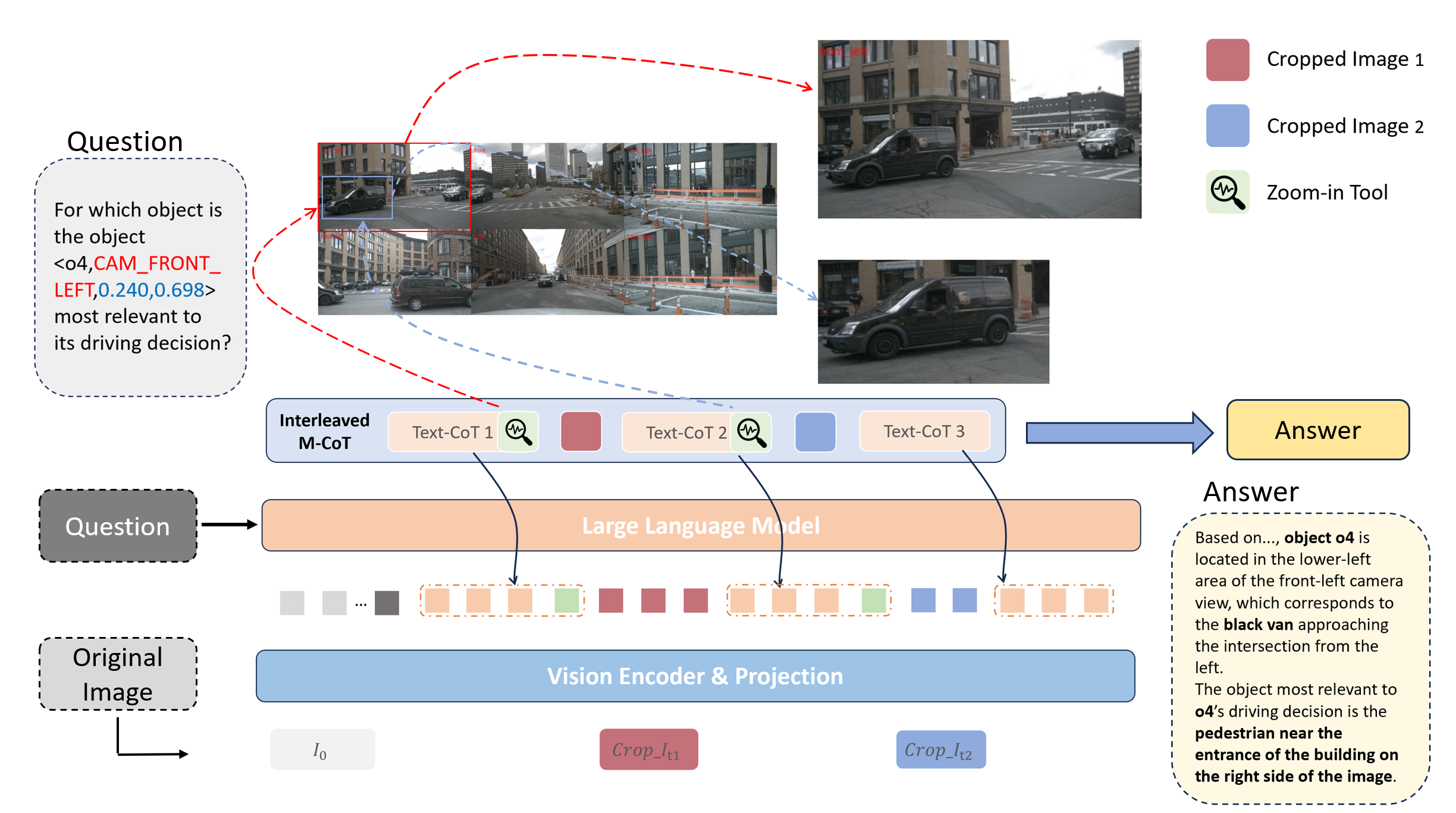}
    \caption{\textbf{The Overall iMCOT Reasoning Framework of OmniDrive-R1.} 
        The model operates in an iterative loop: starting from the \textbf{Original Image ($I_{0}$)} and a \textbf{Question}, the VLM generates a textual thought. It then autonomously decides whether to invoke the \textbf{Zoom-in tool} to actively zoom into a crucial visual region, dynamically acquiring new, fine-grained visual evidence (\textbf{Cropped Image 1 ($I_{1}$)}) based on its native grounding capability. This new input is interleaved into the thought process, allowing the VLM to refine its reasoning iteratively until a reliable answer is produced.}
    \label{fig:fig1}
\end{figure*}

In this section, we first present an overview of our proposed method OmniDirve-R1 in section 3.1. Subsequently, in section 3.2, we detail the two-stage end-to-end RL training procedure along with the corresponding reward design. Finally, we introduce an automated Reinforcement Learning with Verifiable Reward (RLVR) format style data construction pipeline in section 3.3.

\subsection{OmniDrive-R1}
OmniDrive-R1 is a multi-modal driving agent that performs  "thinking with views of different onboard cameras" via an iMCoT reasoning process. The model's grounding and reasoning abilities are jointly optimized through end-to-end reinforcement learning, enabling it to leverage its intrinsic grounding capability to pinpoint crucial task-relevant information.

As depicted in \cref{fig:fig1}, our model takes as input a question $q$ and an original image $I_{0}$ generated from six distinct on-vehicle camera views. To formalize the iterative reasoning process, we define the state $s_t$ of iMCoT at step t as follows:
\begin{equation}
	s_{t} = [(I_{0}, T_{0}), (I_{1}, T_{1}), (I_{t}, T_{t})] = \{I_{\leq t}, T_{\leq t}\},
\end{equation} 
where $I_{\leq t} = \{I_{1},..,I_{t}\}$ represents the cropped image tokens before step t, and $I_{\leq t} = \{I_{1}, ..., I_{t}\}$ 
At each step, OmniDrive-R1 autonomously determines whether to output a final answer or use the zoom-in tool $Tool_{t}$ to acquire information from regions of interest, where $Tool_{t}$ represents the invocation of zoom-in tool at step t. The tool's input comprises a bounding box $b_{t}$ and its corresponding class label $l_{t}$, which are produced by the model in the reasoning step t. Upon successfully invoking the zoom-in tool $Tool_{t}$, the agent returns a crucial region that it identifies as instrumental for the ongoing reasoning process, as exemplified by $I_{t+1}$. Given the state $s_{t}$, action $a_{t}\thicksim \pi_{\theta}(a|s_t)$ is sampled from the policy model $\pi_{\theta}$. This interactive process can be autonomously iterated until a final answer is reached or the maximum number of tool calls is reached. Notably, the text tokens $I_{\leq t}$ and image tokens $T_{\leq t}$ are interleaved in the state. 

\begin{figure*}[t]
    \centering
    \includegraphics[width=\textwidth]{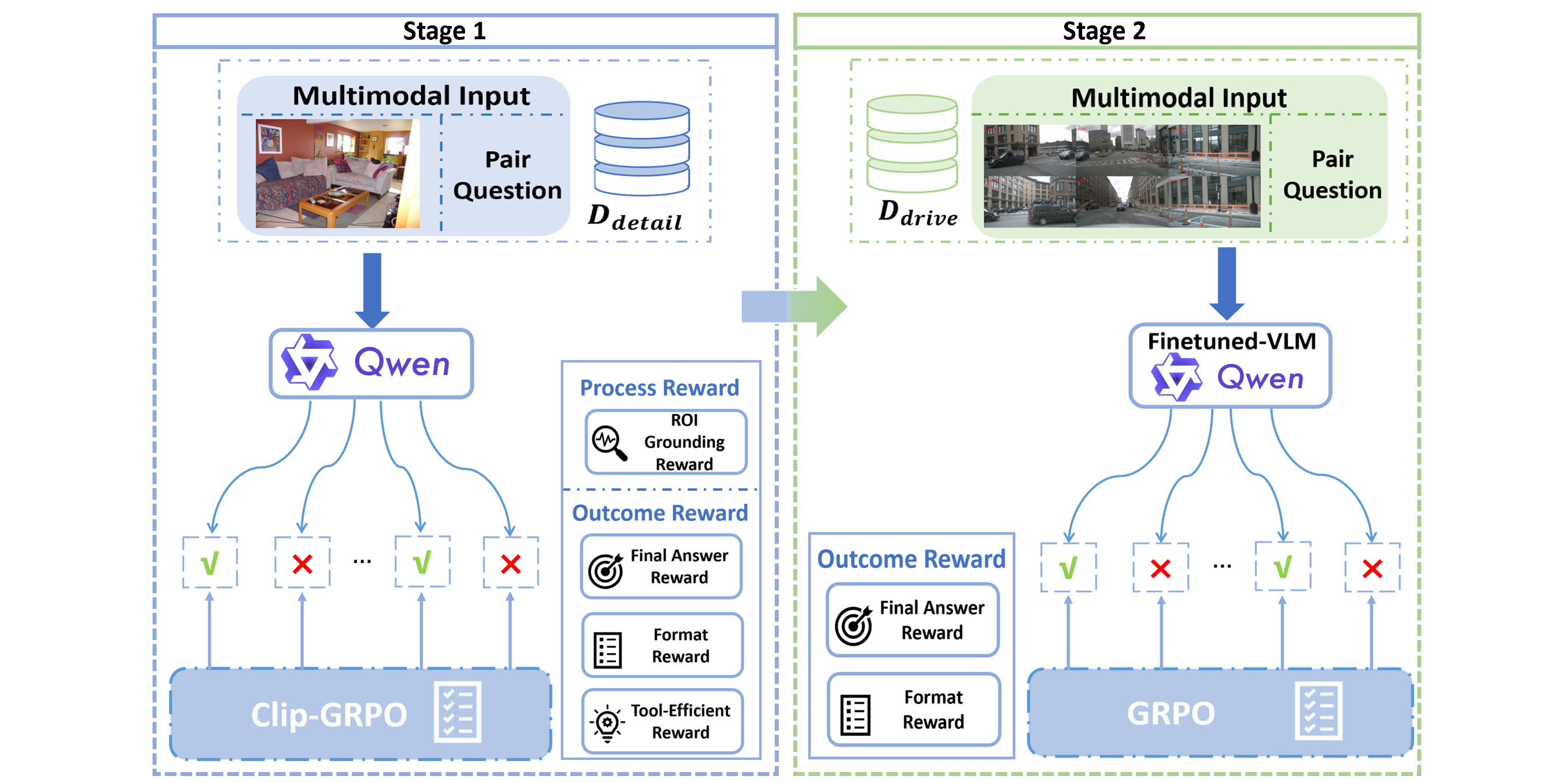}
    \caption{\textbf{The Two-stage Reinforcement Learning Pipeline for OmniDrive-R1.} 
        The training process effectively decouples tool learning from task optimization. \textbf{Stage 1 (Tool Learning, Left)} utilizes the novel Clip-GRPO algorithm on $D_{\text{detail}}$ to enforce robust grounding: the Process Reward (ROI Grounding Reward), which is annotation-free, uses CLIP's cross-modal consistency to ensure the localized region is semantically relevant to the reasoning text. \textbf{Stage 2 (Domain Learning, Right)} fine-tunes the resulting VLM on the autonomous driving dataset ($D_{\text{drive}}$) using GRPO to optimize the strategic timing of tool invocation and the final driving decision (Outcome Reward).}
    \label{fig:fig2}
\end{figure*}

\subsection{Two-stage Training Pipeline}
To enable the model to utilize the zoom-in tool in a context-aware and timely manner, we propose a two-stage reinforcement learning strategy (details provided in \cref{fig:fig2}). Initially, the model is trained on a curated dataset to learn the fundamentals of tool usage. Subsequently, we fine-tune the stage 1 model on autonomous driving datasets to adapt it to real-world driving scenarios and to optimize the timing of tool invocation.

\subsubsection{Stage1: Tool Learning}
To effectively train the model's tool calling capability, we fine-tune it in the first stage on a curated subset $D_{\text{detail}}$ of the DeepEyes dataset \cite{qian2025agentthink}. These selected data points are characterized by a clear correlation between tool usage and improved accuracy, which serves to initially encourage the model to invoke the tool for problem solving.

However, we observed that during inference, the model exhibits a strong bias towards text-based reasoning, leading to infrequent tool invocation. This behavior is particularly pronounced in the early training stages, where the model's intrinsic grounding ability is still nascent. A reward signal based solely on the final outcome can discourage the model from exploring necessary tool usage. Furthermore, while providing human-annotated bounding boxes for critical regions could guide this exploration, it is prohibitively expensive and labor intensive, severely limiting the scalability of our training approach.

Consequently, to foster a cohesive interplay between grounding, reasoning, and tool utilization, we introduce the Clip-GRPO algorithm. This algorithm addresses the aforementioned challenges by proposing a novel reward mechanism, which consists of a process-based region of interest (ROI) grounding reward and an outcome-based tool-efficient reward. 

\textbf{Process-based Reward} We introduce the ROI grounding reward as an instantaneous reward signal to guide the model's grounding behavior upon a successful tool invocation at step t. We leverage a pre-trained CLIP model to compute the similarity score $sim_t$ between the returned region image $I_{t}$ and its corresponding predicted label $l_{t}$ , as shown in Eq.~\eqref{eq:cosine_similarity}:
\begin{equation}
	sim_{t} = \frac{I_t \cdot l_t}{\|I_t\| \cdot \|l_t\|} = \frac{\sum_{k=1}^{D} I_t^k \times l_t^k}{\sqrt{\sum_{k=1}^{D} (I_t^k)^2} \times \sqrt{\sum_{k=1}^{D} (l_t^k)^2}} 
	\label{eq:cosine_similarity},
\end{equation}

This design offers two key advantages: first, it circumvents the need for labor-intensive manual annotation of key regions, allowing for more scalable training. Second, it ensures semantic relevance between the cropped visual content and the generated label, fostering high-quality grounding during the reasoning process. 

And to prevent the model from exploiting this reward by frequently invoking the tool, we introduce a decay coefficient $\lambda$ to mitigate this behavior. This coefficient penalizes excessive tool calls, encouraging the model to use the tool judiciously only when it is genuinely beneficial for solving the task. Therefore, for a reasoning trajectory $\tau$ with $E$ tool calls, the process-based reward is illustrated in Eq.~\eqref{eq:process-based reward}: 
\begin{equation}
	R_{p}(\tau) =  {\sum_{t=1}^{E} sim_t \cdot \lambda^{t-1}}
	\label{eq:process-based reward},
\end{equation}

\textbf{Outcome-based Reward} To reinforce reasoning trajectories that judiciously employ the tool, we adopt a tripartite reward strategy comprising an accuracy reward $R_{acc}$, a formatting reward $R_{f}$, and a reward for the usage of appropriate tools $R_{tool}$. The accuracy reward assesses the correctness of the final answer, while the formatting reward penalizes poorly structured outputs. The tool usage bonus $R_{tool}$ is specifically awarded only when the model achieves a correct final answer and invokes at least one tool during the trajectory. Given a reasoning trajectory $\tau$, outcome-based reward is defined as Eq.~\eqref{eq:outcome-based reward}: 
\begin{equation}
	R_{o}(\tau) =  \alpha R_{acc}(\tau)  + \beta R_{f}(\tau)  + \gamma \mathbb{I}_{R_{acc}(\tau > 0)} R_{tool}
	\label{eq:outcome-based reward},
\end{equation}
where $\alpha$, $\beta$ and $\gamma$ are hyperparameters, $\mathbb{I}_{R_{acc}(\tau > 0)}$ is the indicator function which takes the value of 1 only when $R_{acc}(\tau > 0)$.

In summary, for a reasoning trajectory $\tau$ with $E$ tool calls, the total reward $R(\tau)$ is shown in Eq.~\eqref{eq:total reward}:
\begin{equation}
	R(\tau) = R_{p}(\tau) + R_{o}(\tau) 
	\label{eq:total reward},
\end{equation}

\subsubsection{Stage2: Domain Learning}  
Following the stage 1 fine-tuning, our VLM has acquired robust grounding capabilities and has become proficient at utilizing the tool. Consequently, the primary objective of stage 2 is to enable the model to autonomously decide whether to invoke the tool to capture fine-grained information, based on the complexity of the scene and the query.

To this end, the fine-tuned VLM from stage 1 is optimized using the GRPO algorithm \cite{shao2024deepseekmath} on autonomous driving reasoning datasets, employing a reward strategy that includes accuracy reward $R_{acc}$ and formatting reward $R_f$. Given a reasoning trajectory $\tau$, reward of stage 2 is defined as Eq.~\eqref{eq:stage 2 reward}: 
\begin{equation}
	R(\tau) =  R_{acc}(\tau)  + R_{f}(\tau)
	\label{eq:stage 2 reward},
\end{equation}

\subsection{Data Generation Pipeline}
Prior research has demonstrated that easily verifiable reward signals are crucial for training effective reinforcement learning agents \cite{shao2024deepseekmath}. However, existing autonomous driving datasets, with their open-ended $Q\&A$ format, present significant challenges in terms of reward verification accuracy and scalability.

\begin{figure*}[t]
    \centering
    \includegraphics[width=\textwidth]{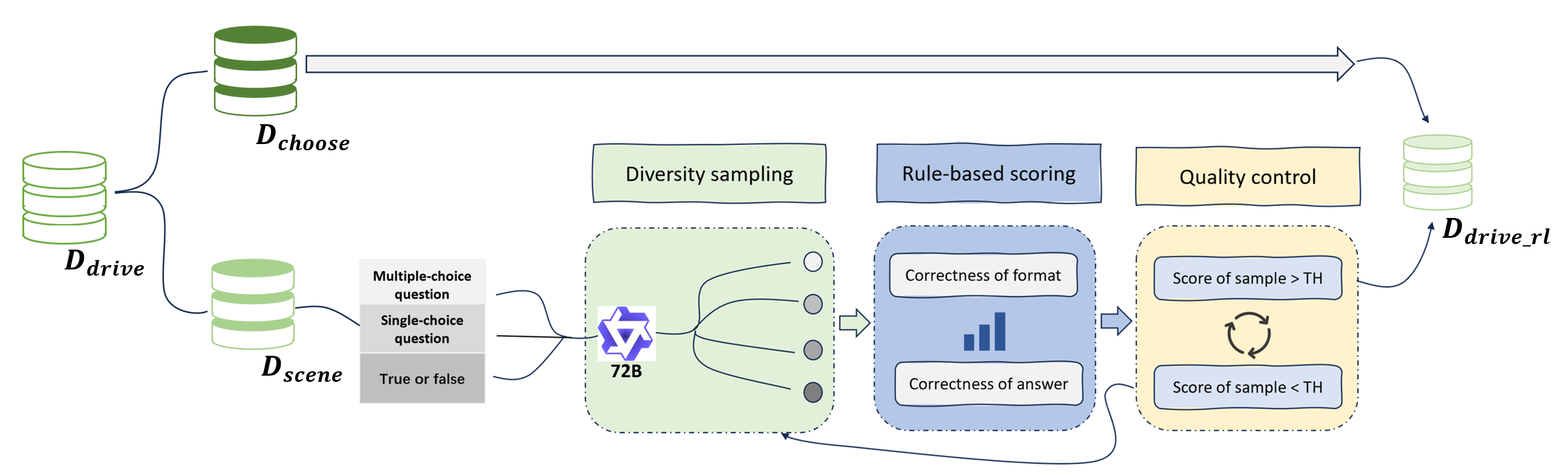}
    \caption{ \textbf{Automated Pipeline for Generating RL-Verifiable Data ($D_{\text{drive\_rl}}$).} 
        To enhance reward verification accuracy and scalability for RL training, open-ended scene $Q\&A$ $D_{\text{scene}}$ from $D_{\text{drive}}$ is converted into structured, easily verifiable formats (Multiple-choice or True/False). The process leverages an advanced MLLM (Qwen2.5VL-72B) for \textbf{Diversity Sampling}, followed by a \textbf{Rule-based Scoring} system (assessing format and answer correctness) and a \textbf{Quality Control} step via rejection sampling to ensure the final dataset $D_{\text{drive\_rl}}$ is high-quality and suitable for efficient RL optimization.}
    \label{fig:fig3}
\end{figure*}

To address this limitation and enhance the utility of these datasets for RL, we introduce a verifiable data generation pipeline, as shown in \cref{fig:fig3}, which converts open-ended scene $Q\&A$ $D_{\text{drive}}$ into easily verifiable multiple-choice or true/false questions. Specifically, we firstly leverage an advanced MLLM, such as Qwen2.5VL-72B \cite{bai2025qwen2}, for diverse candidate sampling. These generated samples are then scored based on a set of predefined rules. A rejection sampling strategy is applied to filter out any samples with scores below a predetermined threshold. Finally, only the highest scoring samples are selected to form our final high-quality dataset $D_{\text{drive\_rl}}$.

%% file: sec/4_exp.tex
\section{Experiment}
\label{sec:exp}
In this section, we first detail our experimental setup, including the datasets, evaluation metrics, and implementation details. Following this, we present a comprehensive comparison of our method against open-source and closed-source VLMs. Furthermore, we conduct ablation studies to validate the effectiveness of each component and generalization tests to assess the robustness of our framework across diverse scenarios.

\subsection{Experiment Setup}
\textbf{Datasets} Our training data is constructed in two stages. For the initial Stage 1: tool learning, we utilize a subset of the DeepEyes dataset \cite{qian2025agentthink}, comprising 14,452 reasoning-based $\text{Q\&A}$ pairs, specifically chosen to encourage tool usage. The subsequent stage 2: domain learning leverages 18,507 $\text{Q\&A}$ pairs sourced from the DriveLMM-o1 dataset \cite{ishaq2025drivelmm}, which is designed to assess a model’s ability to understand perception, prediction, and planning tasks.
In addition, to evaluate our model's performance, we use two benchmarks. We first measure our model's effectiveness on the DriveLLM-o1 evaluation set. We also use the SURDS evaluation set \cite{guo2024surds} for zero-shot testing to assess our model's generalization capability.

\textbf{Evaluation Metrics} We leverage DriveLMM-o1's evaluation metrics. Specifically, we utilize the overall reasoning score to gauge the comprehensive reasoning ability of VLMs, and employ Multiple Choice Quality (MCQ) to assess the accuracy of the final answers. Further details are provided in Appendix A.

\textbf{Implementation Details} We use Qwen2.5VL-7B \cite{bai2025qwen2} as our base model. The training phase is conducted using 16 × NVIDIA A800 GPUs. All RL stages employ GRPO framework with 8 rollouts per question, limiting the maximum number of tool calls to 5 per trajectory.

\subsection{Main Results}
To comprehensively evaluate OmniDrive-R1, we benchmark it against a range of strong open-source VLM models, including InternVL-2.5, LLaVA-CoT, and the Qwen2.5-VL series in a zero-shot setting. We also compare against a suite of state-of-the-art VLM models fintuned on DriveLMM-o1 dataset, such as Agentthink \cite{qian2025agentthink} and DriveLMM-o1 \cite{ishaq2025drivelmm}, as well as closed-source VLMs like GPT-4o. 

The main results are presented in Table~\ref{tab:drivelmm_benchmark}, where our model OmniDrive-R1 achieves the SOTA performance across all categories. Compared to the baseline Qwen2.5VL-7B, we model demonstrates significant improvement, enhancing the overall reasoning score from $51.77\%$ to $80.35\%$ $(+28.58\%)$, and the final answer accuracy from $37.81\%$ to $73.62\%$ $(+35.81\%)$. Compared to the SFT method DriveLMM-o1, OmniDrive-R1 improves the reasoning score by $+5.11\%$ and the MCQ by $+11.26\%$. Furthermore, OmniDrive-R1 surpasses Agentthink in both resaoning ($80.35\%$ vs. $79.68\%$) and MCQ ($73.62\%$ vs. $71.35\%$), demonstrating the efficacy of enhancing a VLM's native grounding and reasoning capabilities to achieve performance superior to methods reliant on extensive, externally-defined toolsets.  

\begin{table*}[t]
\centering
\caption{Evaluation results on the DriveLMM-o1 benchmark. OmniDrive-R1 achieves State-of-the-Art performance across all metrics by leveraging our dynamic Multi-modal Interleaved Chain-of-Thought reasoning.}
\label{tab:drivelmm_benchmark}
\begin{tabular}{l|ccc|cc|cc}
\toprule
\multirow{2}{*}{\textbf{Vision Language Models}} & \multicolumn{3}{c|}{\textbf{Driving Metrics (\%)$\uparrow$}} & \multicolumn{2}{c|}{\textbf{Scene Detail (\%)$\uparrow$}} & \multicolumn{2}{c}{\textbf{Overall(\%)$\uparrow$}} \\
\cline{2-8}
& \textbf{Risk Assess.} & \textbf{Rule Adh.} & \textbf{Scene Aware.} & \textbf{Relevance} & \textbf{Missing} & \textbf{Reason.} & \textbf{MCQ} \\
\midrule
GPT-4o \cite{islam2024gpt}  & 71.32 & 80.72 & 72.96 & 76.65 & 71.43 & 72.52 & 57.84 \\
Ovis1.5-Gemma2-9B \cite{lu2024ovis} & 51.34 & 66.36 & 54.74 & 55.72 & 55.74 & 55.62 & 48.85 \\
Mulberry-7B \cite{yao2024mulberry} & 51.89 & 63.66 & 56.68 & 57.27 & 57.45 & 57.65 & 52.86 \\
LLaVA-CoT \cite{xu2025llava} & 57.62 & 69.01 & 60.84 & 62.72 & 60.67 & 61.41 & 49.27 \\
LlamaV-O1 \cite{thawakar2025llamav}  & 60.20 & 73.52 & 62.67 & 64.66 & 63.41 & 63.13 & 50.02 \\
InternVL2.5-8B \cite{chen2024expanding} & 69.02 & 78.43 & 71.52 & 75.80 & 70.54 & 71.62 & 54.87 \\
Qwen2.5VL-7B \cite{bai2025qwen2} & 46.44 & 60.45 & 51.02 & 50.15 & 52.19 & 51.77 & 37.81 \\
Qwen2.5VL-72B \cite{bai2025qwen2} & 64.40 & 72.81 & 60.29 & 65.13 & 62.81 & 65.73 & 61.27 \\
DriveLMM-o1 \cite{ishaq2025drivelmm} & 73.01 & 81.56 & 75.39 & 79.42 & 74.49 & 75.24 & 62.36 \\
Agentthink \cite{qian2025agentthink} & 80.51 & 84.98 & 82.11 & \textbf{84.99} & \textbf{79.56} & 79.68 & 71.35 \\
\rowcolor{blue!10} % Sets the background color for this entire row
\textbf{OmniDrive-R1 (Ours)} & \textbf{82.31} & \textbf{85.42} & \textbf{83.75} & {82.58} & {78.26} & \textbf{80.35} & \textbf{73.62} \\
\bottomrule
\end{tabular}
\end{table*}

\subsection{Ablation Study}
We conduct a series of ablation experiments to rigorously assess the effectiveness of various reward signals and training strategy within OmniDrive-R1. The quantitative results of these experiments are summarized in Table~\ref{tab:ablation_study_final_clean}.

As observed, ablation studies show that both the progressive two-stage training strategy and structure reward signals yield a significant improvement over the baseline. And the full model (OmniDrive-R1) achieves the SOTA performance across all metrics.

\textbf{Ablation of Training Strategy} We first confirm the necessity of our progressive training approach. We compare the + Clip GRPO model (single-stage training with full reward trained on DriveLMM-o1) with our final OmniDrive-R1 model (two-stage training). The results show that the two-stage approach yields substantial gains in reasoning and final accuracy (Reason: $80.35\%$ vs. $74.83\%$; MCQ: $73.62\%$ vs. $65.48\%$). This $+8.14\%$ improvement in MCQ strongly validates the necessity of the progressive training method for fully unlocking the model's potential.

\textbf{Ablation of Clip-GRPO} Comparing the RL-optimized model (+ SFT, + Clip-GRPO), where both the Supervised Fine-Tuning (SFT) and the Clip-GRPO fine-tuning stages are conducted on DriveLMM-o1, against the baseline sft model (the base model fine-tuned on DriveLMM-o1), shows consistent gains in both reasoning ($76.58\%$ vs. $72.36\%$) and MCQ ($64.38\%$ vs. $62.95\%$). This confirms that the reinforcement learning approach effectively refines the model's policy beyond the limits of sft. Crucially, we contrast our full model (OmniDrive-R1) with the two-stage GRPO method (+ GRPO*), which uses the same training stages with OmniDrive-R1 but omits the process-based grounding reward. The significant performance gap (Reason: $80.35\%$ vs. $70.18\%$; MCQ: $73.62\%$ vs. $57.19\%$) provides compelling evidence that the grounding reward is indispensable for achieving high-quality multi-modal reasoning and guiding judicious tool invocation.

\begin{table*}[h]
\centering
\caption{Ablation study of OmniDrive-R1 on reward design and training strategy. The full model (bottom row) benefits from the combination of Two-stage training strategy, RL, and structured tool-use rewards. Ablation models here trained with 8 epochs.}
\label{tab:ablation_study_final_clean}
\resizebox{\textwidth}{!}{
% 定义列格式：l | cccc | ccc | cc | cc
% SFT Setting 独立，然后是 GRPO Reward Setting，然后是 Driving Metrics, Scene Detail, Overall
\begin{tabular}{l|cc|ccc|ccc|cc|cc}
\toprule
\multirow{2}{*}{\textbf{Model Variant}} & 
\multicolumn{2}{c|}{\textbf{Training Stage}} & 
\multicolumn{3}{c|}{\textbf{GRPO Reward Setting}} & \multicolumn{3}{c|}{\textbf{Driving Metrics (\%)$\uparrow$}} & \multicolumn{2}{c|}{\textbf{Scene Detail (\%)$\uparrow$}} & \multicolumn{2}{c}{\textbf{Overall (\%)$\uparrow$}} \\
\cline{2-13} % 调整 cline 的跨度
& \textbf{Stage1} & \textbf{Stage2} & \textbf{Answer} & \textbf{Tool-eff.} & \textbf{Grounding} & \textbf{Risk Assess.} & \textbf{Rule Adh.} & \textbf{Obj Und.} & \textbf{Relevance} & \textbf{Missing} & \textbf{Reason.} & \textbf{MCQ} \\
\midrule
Base Model & - & - & \myxmark & \myxmark & \myxmark  & 46.44 & 60.45 & 51.02 & 50.15 & 52.19 & 51.77 & 37.81 \\
+ SFT & S & - & \myxmark & \myxmark & \myxmark &  71.13 & 78.65 & 72.31 & 79.27 & 69.32 & 72.36 & 62.95 \\
+ Clip-GRPO & R & - & \mycmark & \mycmark & \mycmark & 75.32 & 77.38 & 74.73 & 75.12 & 74.35 & 74.83 & 65.48 \\
+ GRPO$^{*}$ & R & R & \mycmark & \mycmark & \myxmark & 68.19 & 73.32 & 72.62 & 70.38 & 69.58 & 70.18 & 57.19 \\
+ SFT + Clip-GRPO & S & R & \mycmark & \mycmark & \mycmark & 78.34 & 79.58 & 76.54 & 78.62 & 73.24 & 76.58 & 64.38 \\
\textbf{OmniDrive-R1 (Ours)} & R & R & \mycmark & \mycmark & \mycmark & \textbf{82.31} & \textbf{85.42} & \textbf{83.75} & \textbf{82.58} & \textbf{78.26} & \textbf{80.35} & \textbf{73.62} \\
\bottomrule
\end{tabular}
} % End \resizebox
\end{table*}

\begin{table*}[t]
\centering
% \small 
\caption{Comparison of our proposed method with other open-source and proprietary VLMs, as well as specialized spatial understanding models. Yaw, Pixel, Depth, Dis, L/R, and F/B correspond to the six spatial reasoning tasks. The Score column represents the average performance across these six metrics. \textbf{Bold}: Best. \underline{Underline}: Second Best.}
\label{tab:spatial_comparison}
\begin{tabular}{l|ccc|ccc|c}
\toprule
\multirow{2}{*}{\textbf{Model}} & \multicolumn{3}{c|}{\textbf{Single-object}} & \multicolumn{3}{c|}{\textbf{Multi-object}} & \multirow{2}{*}{\textbf{Score}} \\
\cmidrule(lr){2-4} \cmidrule(lr){5-7}
& \textbf{Yaw} & \textbf{Pixel} & \textbf{Depth} & \textbf{Dis} & \textbf{L/R} & \textbf{F/B} & \\
\midrule
Random & 5.73 & 1.12 & 34.27 & 8.76 & 11.57 & 11.89 & 12.22 \\
\midrule[\heavyrulewidth]
GPT-4o & \underline{13.08} & 1.62 & 2.49 & 11.57 & 47.89 & 3.14 & 13.30 \\
GPT-4o-mini & 3.24 & 0.28 & 0.22 & 4.22 & 21.51 & 2.05 & 5.25 \\
Gemini-1.5-pro \cite{team2024gemini} & \textbf{19.14} & 4.41 & 22.70 & \underline{61.95} & \underline{66.38} & \textbf{22.05} & 32.77 \\
Gemini-2.0-flash & 9.30 & 5.41 & 32.97 & \textbf{69.30} & \textbf{77.30} & \underline{20.00} & \textbf{35.71} \\
LLaVA-OV-Qwen2-72B-si & 1.95 & 3.03 & 23.57 & 3.78 & 9.73 & 8.65 & 8.45 \\
Qwen2.5VL-72B & 11.57 & \underline{6.13} & \textbf{44.00} & 58.05 & 66.16 & 14.92 & \underline{33.47} \\
Qwen2.5VL-7B & 7.57 & 3.46 & {25.95} & 11.46 & 17.95 & 9.30 & 12.61 \\
Qwen2.5-VL-3B & 6.27 & 3.81 & 27.68 & 17.84 & 14.81 & 10.49 & 13.48 \\
\midrule
SpatialBot \cite{cai2025spatialbot} & 0.00 & 0.00 & 12.00 & 0.00 & 0.00 & 0.00 & 2.00 \\
SpatialRGPT \cite{cheng2024spatialrgpt} & 1.30 & 0.55 & 10.59 & 1.95 & 0.86 & 7.35 & 3.77 \\
\midrule
OmniDrive-R1 (Ours) & 9.35 & \textbf{39.46} & \underline{36.72} & 46.25 & 46.51 & 13.42 & {31.95} \\
\bottomrule
\end{tabular}
\end{table*}

\subsection{Generalization Evaluation}
To rigorously assess the robustness and out-of-domain transferability of OmniDrive-R1, we conduct a zero-shot evaluation on the SURDS evaluation set \cite{guo2024surds}. The SURDS benchmark is specifically designed to test fine-grained spatial reasoning, requiring a model to localize and understand objects relative to the ego-vehicle and other scene elements. These tasks are highly sensitive to errors in visual grounding and demand a deep, multi-level understanding of the scene, thus serving as an excellent probe for the model's generalization capabilities. Further details of SURDS are provided in Appendix A.

The benchmark consists of six challenging spatial reasoning tasks: Yaw, Pixel, and Depth for single-object grounding, and Dis (Distance), L/R (Left/Right), and F/B (Front/Back) for multi-object relational reasoning. The results, summarized in Table ~\ref{tab:spatial_comparison}, compare OmniDrive-R1 against a range of leading open-source and proprietary VLMs, as well as specialized spatial understanding models ($\text{SpatialBot}$ \cite{cai2025spatialbot} and $\text{SpatialRGPT}$ \cite{cai2025spatialbot}).

OmniDrive-R1 demonstrates robust zero-shot performance, achieving an overall score of ${31.95}$. This performance is highly competitive, statistically surpassing the baseline Qwen2.5VL-7B-Instruct ($12.61$) and showing comparable capability to large proprietary models such as Gemini1.5-pro ($32.77$) and Qwen2.5-VL72B-Instruct ($33.47$), despite the significant difference in model size.
Notably, our model achieves the best performance in Pixel task (\textbf{$39.46$}). This massive improvement over the $\text{Qwen2.5VL-7B-Instruct}$ baseline ($\text{3.46}$) highlights that the process-based grounding reward effectively refined the VLM's native ability to pinpoint visual regions and establish high-fidelity correspondence between image crops and reasoning text. Additionally, the performance of our model on Depth ($\text{36.72}$) and the multi-object metrics Dis ($\text{46.25}$) and L/R ($\text{46.51}$) also shows a substantial jump compared to the baseline ($\text{Depth: 25.95}$, $\text{Dis: 11.46}$, $\text{L/R: 17.95}$). This consistency across diverse spatial tasks suggests that the iMCoT paradigm fundamentally strengthens the model's spatial reasoning ability, leading to robust generalization on complex relational reasoning tasks.

%% file: sec/5_conclusion.tex
\section{Conclusion}
\label{sec:conclusion}
In this work, we propose OmniDrive-R1 to address the critical challenge of unreliable visual reasoning, especially object hallucination in VLMs for AD. OmniDrive-R1 is the first purely RL–based framework that unifies perception and reasoning through an iMCoT mechanism, enabling the model to dynamically acquire fine-grained visual evidence during reasoning and to leverage its native grounding ability without relying on external detection modules. We further introduced Clip-GRPO, an annotation-free process-level optimization algorithm that uses CLIP-based cross-modal consistency as a grounding reward to jointly enhance region selection and reasoning reliability. And our RLVR data generation pipeline converts open-ended driving questions into verifiable formats, providing a scalable foundation for stable reward computation in open-world driving tasks. Together, these components yield a framework with stronger interpretability, improved grounding fidelity, and superior generalization to previously unseen visual conditions. OmniDrive-R1 achieves SOTA performance on DriveLMM-o1, improving the overall reasoning score from 51.77\% to 80.35\% and MCQ accuracy from 37.81\% to 73.62\%, while also demonstrating competitive zero-shot spatial reasoning on SURDS. 

%% file: sec/X_suppl.tex
\clearpage
\setcounter{page}{1}
\maketitlesupplementary
% 关键步骤 1：加入这行命令，将计数器改为字母 (A, B, C...)
\appendix

\section{Evaluation Metrics Details}
\label{sec:appendix_metrics}

In this section, we provide a detailed explanation of the evaluation metrics used in our experiments, specifically for the DriveLMM-o1 \cite{ishaq2025drivelmm} and SURDS \cite{guo2024surds} benchmarks.

\subsection{DriveLMM-o1 Benchmark Metrics}

Following the evaluation protocol established in previous works~\cite{ishaq2025drivelmm, qian2025agentthink}, we employ a dual-metric system consisting of \textbf{Overall Reasoning Score} and \textbf{Multiple Choice Quality (MCQ)}. To ensure fair comparison and reproducibility, we strictly adhere to the DriveLMM-o1 evaluation protocol. Specifically, we utilize \textbf{GPT-4o-mini} as the automated evaluator. The exact system prompt and the comprehensive scoring criteria employed in our evaluation are presented below:

\begin{enumerate}
    \setlength{\itemsep}{0.5em}

    \item \textbf{Faithfulness-Step (1-10):} Measures how well the model's reasoning steps align with the ground truth.
    \begin{itemize}
        \item \textbf{9-10:} All steps correctly match or closely reflect the reference.
        \item \textbf{7-8:} Most steps align, with minor deviations.
        \item \textbf{5-6:} Some steps align, but several are incorrect or missing.
        \item \textbf{3-4:} Few steps align; most are inaccurate or missing.
        \item \textbf{1-2:} Majority of steps are incorrect.
    \end{itemize}

    \item \textbf{Informativeness-Step (1-10):} Measures completeness of reasoning.
    \begin{itemize}
        \item \textbf{9-10:} Captures almost all critical information.
        \item \textbf{7-8:} Covers most key points, with minor omissions.
        \item \textbf{5-6:} Missing significant details.
        \item \textbf{3-4:} Only partial reasoning present.
        \item \textbf{1-2:} Poor extraction of relevant reasoning.
    \end{itemize}

    \item \textbf{Risk Assessment Accuracy (1-10):} Evaluates if the model correctly prioritizes high-risk objects or scenarios.
    \begin{itemize}
        \item \textbf{9-10:} Correctly identifies and prioritizes key dangers.
        \item \textbf{7-8:} Mostly accurate, with minor misprioritizations.
        \item \textbf{5-6:} Some important risks are overlooked.
        \item \textbf{3-4:} Significant misjudgments in risk prioritization.
        \item \textbf{1-2:} Misidentifies key risks or misses them entirely.
    \end{itemize}

    \item \textbf{Traffic Rule Adherence (1-10):} Evaluates whether the response follows traffic laws and driving best practices.
    \begin{itemize}
        \item \textbf{9-10:} Fully compliant with legal and safe driving practices.
        \item \textbf{7-8:} Minor deviations, but mostly correct.
        \item \textbf{5-6:} Some inaccuracies in legal/safe driving recommendations.
        \item \textbf{3-4:} Several rule violations or unsafe suggestions.
        \item \textbf{1-2:} Promotes highly unsafe driving behavior.
    \end{itemize}

    \item \textbf{Scene Awareness \& Object Understanding (1-10):} Measures how well the response interprets objects, their positions, and actions.
    \begin{itemize}
        \item \textbf{9-10:} Clearly understands all relevant objects and their relationships.
        \item \textbf{7-8:} Minor misinterpretations but mostly correct.
        \item \textbf{5-6:} Some key objects misunderstood or ignored.
        \item \textbf{3-4:} Many errors in object recognition and reasoning.
        \item \textbf{1-2:} Misidentifies or ignores key objects.
    \end{itemize}

    \item \textbf{Repetition-Token (1-10):} Identifies unnecessary repetition in reasoning.
    \begin{itemize}
        \item \textbf{9-10:} No redundancy, very concise.
        \item \textbf{7-8:} Minor repetition but still clear.
        \item \textbf{5-6:} Noticeable redundancy.
        \item \textbf{3-4:} Frequent repetition that disrupts reasoning.
        \item \textbf{1-2:} Excessive redundancy, making reasoning unclear.
    \end{itemize}

    \item \textbf{Hallucination (1-10):} Detects irrelevant or invented reasoning steps not aligned with ground truth.
    \begin{itemize}
        \item \textbf{9-10:} No hallucinations, all reasoning is grounded.
        \item \textbf{7-8:} One or two minor hallucinations.
        \item \textbf{5-6:} Some fabricated details.
        \item \textbf{3-4:} Frequent hallucinations.
        \item \textbf{1-2:} Majority of reasoning is hallucinated.
    \end{itemize}

    \item \textbf{Semantic Coverage-Step (1-10):} Checks if the response fully covers the critical reasoning elements.
    \begin{itemize}
        \item \textbf{9-10:} Nearly complete semantic coverage.
        \item \textbf{7-8:} Good coverage, some minor omissions.
        \item \textbf{5-6:} Partial coverage with key gaps.
        \item \textbf{3-4:} Major gaps in reasoning.
        \item \textbf{1-2:} Very poor semantic coverage.
    \end{itemize}

    \item \textbf{Commonsense Reasoning (1-10):} Assesses the use of intuitive driving logic in reasoning.
    \begin{itemize}
        \item \textbf{9-10:} Displays strong commonsense understanding.
        \item \textbf{7-8:} Mostly correct, with minor gaps.
        \item \textbf{5-6:} Some commonsense errors.
        \item \textbf{3-4:} Frequent commonsense mistakes.
        \item \textbf{1-2:} Lacks basic driving commonsense.
    \end{itemize}

    \item \textbf{Missing Step (1-10):} Evaluates if any necessary reasoning steps are missing.
    \begin{itemize}
        \item \textbf{9-10:} No critical steps missing.
        \item \textbf{7-8:} Minor missing steps, but answer is mostly intact.
        \item \textbf{5-6:} Some important steps missing.
        \item \textbf{3-4:} Many critical reasoning gaps.
        \item \textbf{1-2:} Response is highly incomplete.
    \end{itemize}

    \item \textbf{Relevance (1-10):} Measures how well the response is specific to the given scenario and ground truth.
    \begin{itemize}
        \item \textbf{9-10:} Highly specific and directly relevant to the driving scenario. Captures critical elements precisely, with no unnecessary generalization.
        \item \textbf{7-8:} Mostly relevant, but some minor parts may be overly generic or slightly off-focus.
        \item \textbf{5-6:} Somewhat relevant but lacks precision; response contains vague or general reasoning without clear scenario-based details.
        \item \textbf{3-4:} Mostly generic or off-topic reasoning, with significant irrelevant content.
        \item \textbf{1-2:} Largely irrelevant, missing key aspects of the scenario and failing to align with the ground truth.
    \end{itemize}

    \item \textbf{Missing Details (1-10):} Evaluates the extent to which critical information is missing from the response, impacting the reasoning quality.
    \begin{itemize}
        \item \textbf{9-10:} No significant details are missing; response is comprehensive and complete.
        \item \textbf{7-8:} Covers most important details, with minor omissions that do not severely impact reasoning.
        \item \textbf{5-6:} Some essential details are missing, affecting the completeness of reasoning.
        \item \textbf{3-4:} Many critical
        reasoning steps or contextual details
        are absent, making the response incomplete.
        \item \textbf{1-2:} Response is highly lacking in necessary details, leaving major gaps in understanding.
    \end{itemize}
\end{enumerate}

\subsection{SURDS Benchmark Metrics}
\label{subsec:surds_metrics}

The SURDS benchmark~\cite{guo2024surds} is designed to rigorously evaluate the fine-grained spatial reasoning capabilities of VLMs. It comprises six distinct tasks divided into single-object and multi-object categories. We report the \textbf{accuracy (\%)} for each task, utilizing specific calculation protocols based on the output type.

\noindent \textbf{Single-object Spatial Reasoning:}
\begin{itemize}
    \item \textbf{Yaw (Orientation):} Evaluates the model's ability to estimate the heading angle of a specific object. The metric adopts a classification accuracy based on normalized textual matching.
    
    \item \textbf{Pixel (Localization):} Tests the model's capability to ground a textual description to a specific 2D region. Unlike standard bounding box IoU, SURDS employs a centerness score to evaluate the precision of the predicted point $P=(x,y)$ relative to the ground truth bounding box $B=\{x_{min}, y_{min}, x_{max}, y_{max}\}$.
    
    If $P$ falls outside $B$, the score is 0. If inside, we compute the distances to the four borders ($l, r, t, b$) and the center-aligned ratios:
    \begin{equation}
        \mathcal{R}_{lr} = \frac{\min(l, r)}{\max(l, r)}, \quad \mathcal{R}_{tb} = \frac{\min(t, b)}{\max(t, b)}
    \end{equation}
    The score for a sample is defined as the geometric mean of these ratios: $S_{pixel} = \sqrt{\mathcal{R}_{lr} \cdot \mathcal{R}_{tb}}$.
    
    \item \textbf{Depth:} Assesses the model's ability to estimate the absolute distance or depth range. Similar to Yaw, this is evaluated via textual classification accuracy.
\end{itemize}

\noindent \textbf{Multi-object Relational Reasoning:}
\begin{itemize}
    \item \textbf{Dis (Distance):} Evaluates the understanding of pairwise distances (e.g., determining if Object A is closer than Object B).
    \item \textbf{L/R (Left/Right):} Tests the ability to resolve lateral spatial relationships between two objects.
    \item \textbf{F/B (Front/Back):} Measures the comprehension of longitudinal spatial relationships in 3D space.
\end{itemize}

\noindent \textbf{Metric for Classification Tasks:}
For the five tasks excluding Pixel (i.e., Yaw, Depth, Dis, L/R, F/B), we employ a normalized exact match accuracy. To ensure robust evaluation, we define a normalization function $\mathcal{N}(\cdot)$ which removes punctuation, articles (e.g., "a", "the"), and extra whitespace, converting text to lowercase. A prediction $A_{pred}$ is considered correct if and only if:
\begin{equation}
    \mathcal{N}(A_{pred}) = \mathcal{N}(A_{gt})
\end{equation}
The \textbf{Overall Score} for SURDS is calculated as the arithmetic mean of the accuracy scores across these six tasks, providing a holistic view of the VLM's spatial intelligence.